\documentclass[letterpaper, 10 pt, conference]{ieeeconf}  

\IEEEoverridecommandlockouts                              

\usepackage[table]{xcolor}
\makeatletter
\let\labelindent\@undefined
\makeatother
\usepackage{enumitem}
\usepackage{amsmath}
\usepackage{amssymb}
\usepackage{amsfonts}
\usepackage{bbm}
\usepackage{bm}
\usepackage{algorithm}
\usepackage{algpseudocode}
\usepackage{float}  

\usepackage[skip=1pt, font=small]{caption}

\usepackage{booktabs}
\usepackage{array}
\usepackage{multirow}
\usepackage{makecell}
\usepackage{graphicx}   
\usepackage{rotating}
\usepackage{wrapfig}
\usepackage{subcaption}
\definecolor{baselinerow}{HTML}{ECECEC}  
\definecolor{ablationrow}{HTML}{FFF6D6}  
\definecolor{proposedrow}{HTML}{DDF2DD}  

\usepackage{hyperref} 

\title{\LARGE \bf
Non-Prehensile Multi-Object Placement with Mobile Robots \\ via Diffusion-Based Trajectory Planning
}

\author{Lance Pharand$^{1}$, Megnath Ramesh, Mark Chiu, Yash Vardhan Pant, Stephen L. Smith
\thanks{$^{1}$ This research is supported in part by the Natural Sciences and Engineering Research Council of Canada (NSERC) and the Canada Research Chairs Program. This research was enabled in part by support provided by the Vector Institute (https://vectorinstitute.ai) and the Digital Research Alliance of Canada (alliance.can.ca). Lance Pharand is also financially supported by the Vector Scholarship in Artificial Intelligence, provided through the Vector Institute. All authors are part of the Electrical and Computer Engineering Department at the University of Waterloo, ON, Canada}%
}

\begin{document}

\maketitle
\thispagestyle{empty}
\pagestyle{empty}



    
\begin{abstract}
    Non-prehensile manipulation with mobile robots has primarily targeted placement in coarse goal regions. Precision placement has been studied with manipulators on benchmarks such as Push-T, where an omnidirectional end effector can readily re-approach an object from any side. For a mobile robot, correcting the same placement errors may require wide repositioning maneuvers around the object. We formalize multi-object pushing with a mobile robot as a multi-target assignment problem at tight tolerances. We propose a diffusion planner with a causal factorization: each robot denoising step is followed by an object step conditioned on the current clean robot estimate. This factorization allows us to sample multiple object outcomes for a single robot plan, capturing the multi-modality of pushing. We also design a set of pushing-specific costs for mobile robots, used for both guidance and best-of-$N$ sampling at inference. On Push-T, our planner attains the highest success rate of the methods we compare, while having the lowest rate of implausible object motion. Across three multi-box pushing environments resembling warehouse scenarios, it raises the fraction of boxes placed by $21.5$ percentage points over an unguided baseline, and replans $1.8\times$ faster than a planner that denoises the two trajectories together. Lastly, we deploy the guided planner on a TurtleBot 3 Burger with object shapes absent from the training data, where it places $85.0\%$ of objects within tolerance against Diffusion Policy's $37.5\%$.
\end{abstract}

\section{Introduction}
    
\begin{figure}
\centering
\includegraphics[width=0.7\columnwidth]{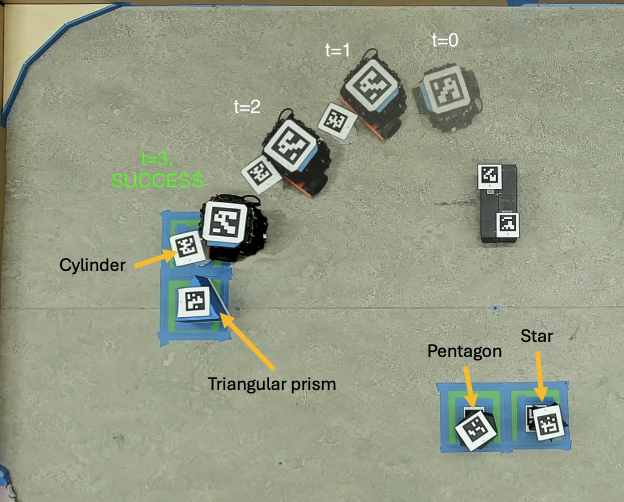}
\caption{Robot completes a successful push during an Out-of-distribution (OOD) experiment with new shapes. Robot demo videos and code available on \href{https://anonymous.4open.science/r/Causal_Diffusion_Planner-0181}{our project website}
}
\label{fig:robot_demo}
\vspace{-0.5cm}
\end{figure}

In this work, we focus on the problem of multi-object pushing with a mobile robot. Formally, this is a multi-target assignment problem, where each object must be placed inside a tight target configuration (Fig.~\ref{fig:robot_demo}). Successfully solving this problem would make mobile robots viable for warehouse tasks where they must push boxes or pallets to precise positions for downstream automation to take over. Precision pushing has been studied in the manipulator setting, where Push-T \cite{chiDiffusionPolicyVisuomotor2024, florenceImplicitBehavioralCloning2021} is a popular benchmark \cite{jiangStreamingFlowPolicy2025, panModelBasedDiffusionTrajectory}. Here, a robot with a circular end-effector must push a T-shaped block onto a goal pose, with success requiring near-complete overlap. In contrast, existing non-prehensile manipulation (NPM) work with mobile robots~\cite{wuSpatialActionMaps, zhongBenchPushBenchmarkingPushingbased2025, caroPushSmarterNot2025a} only requires the object to land inside a target zone much larger than the object itself.

Two properties of the mobile robot setting make pushing an object to within a small tolerance of a target harder than it is in the manipulator setting. First, mobile robots typically have non-holonomic dynamics and cannot move sideways. Second, the robot's contact face is typically a non-prehensile bumper fixed to the front, so changing the side of the object being pushed requires a wide reposition. Both aspects make it difficult to perform the small corrections needed to bring an object within a tight tolerance of its target. In contrast, the end effector in Push-T can move sideways, is omnidirectional, and can re-approach the object from any side. Planners and policies for the studied setting must account for the difficulty of these corrections, but existing mobile robot NPM methods are evaluated in settings where this is not necessary.




Diffusion-based planners have shown promise in solving difficult pushing problems \cite{jiangStreamingFlowPolicy2025, panModelBasedDiffusionTrajectory}. Diffusion planners that predict object motion typically denoise the robot and object trajectories simultaneously~\cite{luoGenerativeTrajectoryStitching2025}, which can yield object motion that is implausible given the robot's actions. We instead factorize the joint trajectory distribution into a robot trajectory prior and an object trajectory conditional. Each robot denoising step is followed by an object step conditioned on the current clean robot estimate (Fig.~\ref{fig:causal}). As a result, guidance can be evaluated on a predicted object outcome that is a response to the current robot plan, and its gradient is backpropagated through that response to improve the robot plan. The Causal factorization also lets us sample several object outcomes for a single robot plan, which is beneficial as pushing is multimodal~\cite{7989345}. Our \textbf{main contributions} are:

\begin{enumerate}[itemsep=2pt, topsep=2pt, leftmargin=*]
    \item We formulate and evaluate multi-object placement by a mobile robot under tight spatial tolerances, requiring each object's footprint to lie within its target region and previously completed placements to remain undisturbed. We assess how placement tolerance affects execution success across multiple environments.
    \item We propose a diffusion-based planner that uses a Causal factorization of trajectory denoising for non-prehensile pushing. With this, each robot denoising step is followed by an object step that is conditioned on the current clean robot estimate. We also propose a guidance formulation for this factorization, together with a set of differentiable cost terms tailored to mobile robot pushing. 
\end{enumerate}

In this work, our evaluation spans the Push-T task, three unseen mobile robot pushing environments, and a hardware deployment on a TurtleBot 3 Burger, with analysis of the planner's capabilities and limitations.

\subsection{Related Works}



\paragraph{Diffusion models for planning} Diffusion-based trajectory planners have been applied to both short and long-horizon tasks, both of which arise in NPM. At short horizons, the planner samples a trajectory, executes a portion of it, and then replans \cite{chiDiffusionPolicyVisuomotor2024, jiangStreamingFlowPolicy2025}. Long horizon settings instead tend to rely on trajectory stitching, where a full plan is composed from sub-trajectories already present in the training distribution \cite{luoGenerativeTrajectoryStitching2025, clarkWhatYouNeed2026}. Closest to our proposed factorization is MBA~\cite{suMotionActionDiffusing}, which also splits robot and object denoising into two phases. However, MBA fully denoises the object trajectory first and conditions the robot denoising on that finished prediction, whereas our phases alternate within each denoising step. Diffusion Forcing~\cite{chen2024diffusion} also orders denoising, but through per-token noise levels rather than separate phases, resolving lower-noise tokens first. Unlike our method, most of these planners do not apply guidance at inference, which can limit their ability to adapt to task settings outside their training distribution at deployment.

\paragraph{Guidance in diffusion models} Guidance is applied to diffusion models across many domains such as image generation \cite{dhariwalDiffusionModelsBeat2021, hoClassifierFreeDiffusionGuidance2022, epsteinDiffusionSelfGuidanceControllable2023}, autonomous driving, and robotics. Diffuser \cite{jannerPlanningDiffusionFlexible2022a} steers plans by nudging the denoising process along the gradient of a learned return predictor. However, the performance of this learned model on novel scenes is dependent on its training distribution. Diffusion-ES \cite{yangDiffusionESGradientfreePlanning2024} skips gradient-based guidance entirely. Instead, it samples trajectories from an unconditional diffusion model, scores them with a reward function, and projects elites back onto the data manifold. This keeps the trajectory distribution broader and improves robustness, but the search across iterations is expensive. Diffusion Planner \cite{zhengDiffusionBasedPlanningAutonomous2025} is a transformer-based driving planner that relies on hand-designed energy functions at inference, eliminating the need for an auxiliary model. Our work builds upon this concept in two ways. First, we extend these energy functions to have asymmetric penalties for precision pushing tasks. Second, we add a per-term gradient rescaling that lets each term's contribution be tuned independent of its native scale and units.

\paragraph{Pushing tasks} Diffusion-based trajectory planners have been applied to pushing in non-precision settings, beyond just Push-T. CompDiffuser \cite{luoGenerativeTrajectoryStitching2025} reports encouraging results on AntSoccer, where a quadruped agent pushes a ball to a goal location, and HeRD \cite{caroPushSmarterNot2025a} demonstrates similar gains in a multi-object box environment. Precision has instead been pursued through reinforcement learning by \cite{bergmannPrecisionFocusedReinforcementLearning2024a}, who train a recurrent policy that pushes single objects to within 1 cm of the goal. To the best of our knowledge, no diffusion-based planner targets multi-object NPM with a mobile robot at tight tolerances.

\section{Preliminaries}

\label{sec:prelims}

\subsection{Diffusion models and DDIM sampling}
\label{sec:prelims:diffusion}
A denoising diffusion probabilistic model (DDPM) \cite{hoDenoisingDiffusionProbabilistic2020} defines a forward Markov chain that corrupts a clean sample $\bm{x}_0 \sim p_{\text{data}}$ with Gaussian noise according to a variance schedule $\{\beta_t\}_{t=1}^T$. With $\bar{\alpha}_t = \prod_{s=1}^t (1 - \beta_s)$, the noisy sample $\bm{x}_t$ can be obtained directly from $\bm{x}_0$ in a single step as $\bm{x}_t = \sqrt{\bar{\alpha}_t}\, \bm{x}_0 + \sqrt{1 - \bar{\alpha}_t}\, \bm{\epsilon}$, $\bm{\epsilon} \sim \mathcal{N}(\bm{0}, \bm{I})$. A neural network $\bm{\epsilon}_\theta(\bm{x}_t, t)$ is trained to predict $\bm{\epsilon}$ via the objective $\mathbb{E}_{t, \bm{x}_0, \bm{\epsilon}}\|\bm{\epsilon}_\theta(\bm{x}_t, t) - \bm{\epsilon}\|^2$ and the network recovers the score function directly as $\nabla_{\bm{x}_t}\!\log p(\bm{x}_t) = -\bm{\epsilon}_\theta(\bm{x}_t, t) / \sqrt{1 - \bar{\alpha}_t}$ \cite{luoUnderstandingDiffusionModels2022}. We sample with DDIM \cite{songDenoisingDiffusionImplicit2022a} at $\eta = 0$, a deterministic reverse process derived from a non-Markovian forward formulation that reuses the trained $\bm{\epsilon}_\theta$. The update from $\bm{x}_t$ to $\bm{x}_{t-1}$ depends on a clean estimate $\hat{\bm{x}}_0(\bm{x}_t)$ of the sample: $\bm{x}_{t-1} = \sqrt{\bar{\alpha}_{t-1}}\, \hat{\bm{x}}_0(\bm{x}_t) + \sqrt{1 - \bar{\alpha}_{t-1}}\, \bm{\epsilon}_\theta(\bm{x}_t, t)$ where $\hat{\bm{x}}_0(\bm{x}_t)$ is given by Tweedie's formula \cite{efronTweediesFormulaSelection2011},
\begin{equation}
\label{eq:tweedie}
\hat{\bm{x}}_0(\bm{x}_t) \;=\; \mathbb{E}\!\left[\bm{x}_0 \mid \bm{x}_t\right]
\;=\; \frac{\bm{x}_t - \sqrt{1 - \bar{\alpha}_t}\, \bm{\epsilon}_\theta(\bm{x}_t, t)}{\sqrt{\bar{\alpha}_t}}
\end{equation}

Under $\epsilon$-prediction, this estimate divides by $\sqrt{\bar{\alpha}_t}$ and is ill-conditioned as $\bar{\alpha}_t \to 0$ (Eq.~\ref{eq:tweedie}). We therefore use $v$-prediction~\cite{salimans2022progressive}, where the network predicts $\bm{v}_t = \sqrt{\bar{\alpha}_t}\,\bm{\epsilon} - \sqrt{1-\bar{\alpha}_t}\,\bm{x}_0$ and the Tweedie estimate $\hat{\bm{x}}_0(\bm{x}_t) = \sqrt{\bar{\alpha}_t}\,\bm{x}_t - \sqrt{1-\bar{\alpha}_t}\,\bm{v}_\theta(\bm{x}_t, t)$ involves no division.



\subsection{Gradient-based guidance: DPS}
\label{sec:prelims:guidance}
Test time guidance steers a trained diffusion model toward samples consistent with an auxiliary signal $\mathbf{y}$ without retraining. Following~\cite{jannerPlanningDiffusionFlexible2022a, zhengDiffusionBasedPlanningAutonomous2025}, let $\mathbf{y} \in \{0,1\}$ be an indicator variable with likelihood $p(\mathbf{y}{=}1 \mid \bm{x}_0) \propto e^{-\mathcal{C}(\bm{x}_0)}$, where $\mathcal{C}(\bm{x}_0)$ is a cost associated with $\bm{x}_0$. The reverse process should then sample from the conditional $p(\bm{x}_t \mid \mathbf{y})$ rather than the unconditional prior. The corresponding posterior score decomposes by Bayes' rule as $\nabla_{\bm{x}_t}\!\log p(\bm{x}_t \mid \mathbf{y}) = \nabla_{\bm{x}_t}\!\log p(\bm{x}_t) + \nabla_{\bm{x}_t}\!\log p(\mathbf{y} \mid \bm{x}_t)$. The first term is the prior score, given by the trained $\bm{\epsilon}_\theta$ (Sec.~\ref{sec:prelims:diffusion}). The second is the guidance term, which is intractable because $\mathbf{y}$ depends on the clean $\bm{x}_0$ rather than the noised $\bm{x}_t$. We therefore require an approximation of the guidance term to estimate the posterior score. For this, we use diffusion posterior sampling (DPS) \cite{chungDiffusionPosteriorSampling2024} which can directly estimate the guidance expectation $\mathbb{E}_{\bm{x}_0 \sim p(\bm{x}_0 \mid \bm{x}_t)}[p(\mathbf{y} \mid \bm{x}_0)]$. DPS replaces it with the point estimate $p(\mathbf{y} \mid \hat{\bm{x}}_0(\bm{x}_t))$ by evaluating at the Tweedie posterior mean ({Eq.~\eqref{eq:tweedie}}). As a result, the guidance term becomes $\nabla_{\bm{x}_t}\!\log p(\mathbf{y} \mid \hat{\bm{x}}_0(\bm{x}_t))$, which is tractable. 


\section{Problem formulation}
\label{sec:prelims:problem}
Our setup for the multi-object mobile robot pushing setting follows prior NPM work \cite{caroPushSmarterNot2025a}. The robot is equipped with a non-prehensile bumper, so it can only push, and an object moves only while it is in contact with the bumper. Consider a closed planar workspace $\mathcal{W} \subseteq \mathbb{R}^2$ containing $M$ movable objects, $N$ fixed obstacles $\mathcal{O}_1, \dots, \mathcal{O}_N \subseteq \mathcal{W}$, and a mobile robot with pose $\mathbf{r}(t) \in \mathrm{SE}(2)$. Let $b_m(t) \subseteq \mathcal{W}$ be the region occupied by object $m \in \{1,\dots,M\}$ at time $t$, and let $\mathcal{T}_m \subseteq \mathcal{W}$ be its target configuration.\footnote{We use \emph{target configuration} rather than \emph{goal} because the latter typically denotes a single agent's destination, whereas our task requires defining a target region for each of the $M$ objects.} Each target $\mathcal{T}_m$ is a square with a margin $\delta \geq 0$ around the object's footprint, where a smaller $\delta$ tightens the tolerance. Object $m$ is placed at time $t'$ when $b_m(t') \subseteq \mathcal{T}_m$ and the robot is no longer in contact with object $m$. The task is to place every object within the time budget $T_{\max}$ while avoiding all placed objects. The robot navigates the free space $\mathcal{F}(t) = \mathcal{W} \setminus \big( \bigcup_{i=1}^{N} \mathcal{O}_i \cup \bigcup_{m \in \mathcal{P}(t)} b_m(t) \big)$, where $\mathcal{P}(t) \subseteq \{1,\dots,M\}$ indexes the placed objects at time $t$. We decompose the NPM problem into $M$ subproblems by solving a traveling salesman problem (TSP) that assigns boxes to targets and orders them for execution. For $t < T_{\max}$, a trajectory planner predicts trajectories of length $T_p$. Some planners predict a robot trajectory $\bm{\tau}^R$ paired with an object trajectory $\bm{\tau}^O$ whereas others only predict $\bm{\tau}^R$. At step $k \in \{0,\dots,T_p-1\}$, the predicted robot pose is $\bm{\tau}^R_k \in \mathrm{SE}(2)$ and the predicted object position is $\bm{\tau}^O_k \in \mathbb{R}^2$. The robot executes $\bm{\tau}^R$, after which the trajectory planner replans.

\section{Method}
\label{sec:method}
In this section, we first describe the Causal factorization diffusion model and its main comparison, the Joint prediction model. Following this, we introduce the guidance formulation and the cost function that penalizes undesirable overshoots and safety infractions during pushing. 


\subsection{Diffusion model architectures}
\label{sec:method:arch}

\begin{figure*}
\centering
\includegraphics[width=0.98\textwidth]{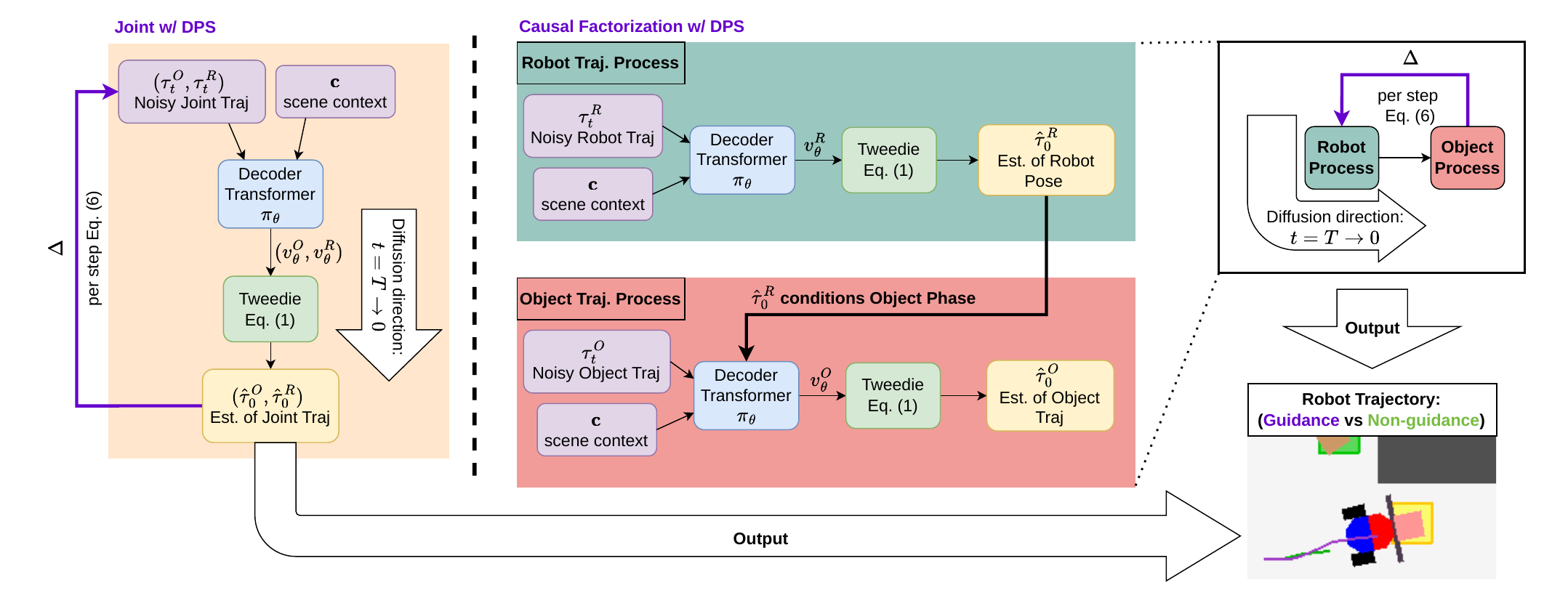}
\caption{Architecture diagrams of the Joint prediction and Causal factorization planners. The Joint planner denoises both robot and object trajectories together. In contrast, the Causal planner makes two passes through a shared transformer decoder. The Robot phase runs first and Tweedie's formula (Eq.~\eqref{eq:tweedie}) turns this output into a clean robot estimate $\hat{\bm{\tau}}^R_0$. The Object phase denoises $\bm{\tau}^O_t$ while being conditioned on $\hat{\bm{\tau}}^R_0$. The resulting chain $\bm{\tau}^R_t \to \hat{\bm{\tau}}^R_0 \to \hat{\bm{\tau}}^O_0$ is differentiable through both decoder passes.}
\label{fig:causal}
\vspace{-0.5cm}
\end{figure*}

\textbf{Causal factorization model.} In pushing, the robot's motion determines the object's. We encode this dependency through the factorization $p(\bm{\tau}^R, \bm{\tau}^O \mid \mathbf{c}) = p(\bm{\tau}^R \mid \mathbf{c})\, p(\bm{\tau}^O \mid \bm{\tau}^R, \mathbf{c})$. We refer to the two factors as the \emph{Robot phase} and \emph{Object phase} (Fig.~\ref{fig:causal}), and both phases share a single transformer decoder.

The Object phase conditions on the clean Tweedie estimate $\hat{\bm{\tau}}^R_0$ produced by the Robot phase. This dependence drives two design choices, since $\hat{\bm{\tau}}^R_0$ is unreliable early in the reverse process. First, we train with the $v$-parameterization~\cite{salimans2022progressive}, which keeps the estimate bounded as $\bar{\alpha}_t \to 0$. Second, $\hat{\bm{\tau}}^R_0$ is close to pure noise early in the reverse process, so we gate the robot conditioning tokens using the signal fraction of the diffusion schedule, $\mathbf{h}^R_t = \bar{\alpha}_t\, \phi(\hat{\bm{\tau}}^R_0) + (1 - \bar{\alpha}_t)\, \mathbf{h}_{\emptyset}$, where $\phi$ is an embedding space projection and $\mathbf{h}_{\emptyset}$ is a learned null embedding. The intention is for the Object phase to fall back to a marginal prediction when the robot plan carries no usable signal, and recover the causal conditional $p(\bm{\tau}^O \mid \bm{\tau}^R, \mathbf{c})$ as the robot trajectory denoises.

\textbf{Joint prediction model.} We also compare with a joint trajectory denoiser, which models $p(\bm{\tau}^R, \bm{\tau}^O \mid \mathbf{c})$. This model is not considered causal under our definition since it does not condition the object denoising on a clean robot trajectory $\hat{\bm{\tau}}^R_0$. As a result, Eq.~\ref{eq:guidance-grad-chain}'s two rightmost factors fuse into one, $\partial \hat{\bm{\tau}}^O_0 / \partial \bm{\tau}^R_t$. 



\textbf{Network.} For both the Causal and Joint models, we use a decoder composed of a stack of $N$ DiT-style transformer blocks~\cite{peeblesScalableDiffusionModels2023, zhengDiffusionBasedPlanningAutonomous2025}. Following the divided space-time attention scheme of~\cite{bertasiusSpaceTimeAttentionAll2021}, each block applies temporal self-attention separately along each trajectory, then inter-trajectory self-attention between the robot and object trajectories at each timestep. For the Causal model, a mask on the inter-trajectory attention prevents clean robot tokens from attending to the noisy object, so the robot estimate stays noise-free. Fig.~\ref{fig:causal} displays the trajectory generation process for a single denoising step.

\subsection{Goal-conditioned guidance}
\label{sec:method:guidance}
For mobile robot pushing tasks, we train the planners to imitate expert demonstrations that avoid both overshooting placement and poor pushing approaches. Despite this, the planner still overshoots at inference, as the denoising objective never penalizes it (Sec.~\ref{mobile_robot_sec}). To penalize overshoots, we design a cost function $\mathcal{C}$ over predicted trajectories. It is composed of multiple terms (Sec.~\ref{sec:method:cost}), each penalizing a distinct undesirable outcome in pushing such as overshoot, collision, or a poor initial approach. A key property of the Causal design is the chain in Eq.~\ref{eq:guidance-grad-chain}, where the gradient of the guidance function is allowed to flow back to the noisy robot trajectory. The result is more successful placements (Sec.~\ref{sec:experiments}).


\textbf{Setup.} Following Sec.~\ref{sec:prelims:guidance}, we condition the reverse diffusion process on the goal-reached event $\mathbf{y}$ with likelihood $p(\mathbf{y} \mid \bm{\tau}^O_0) \propto \exp(-\mathcal{C}(\bm{\tau}^O_0) / \lambda)$ where $\lambda$ is the Boltzmann temperature~\cite{jannerPlanningDiffusionFlexible2022a, zhengDiffusionBasedPlanningAutonomous2025}. For both planners, $\mathbf{y}$ depends on the clean object trajectory $\bm{\tau}^O_0$ rather than on $\bm{\tau}^R_t$ directly, so the guidance term $\nabla_{\bm{\tau}^R_t}\!\log p(\mathbf{y} \mid \bm{\tau}^R_t)$ is intractable. We hereafter write $p(\mathbf{y} \mid \cdot)$ as shorthand for $p(\mathbf{y}{=}1 \mid \cdot)$.

\textbf{A differentiable guidance chain.} In the Causal planner, the Robot phase maps $\bm{\tau}^R_t$ to a clean estimate $\hat{\bm{\tau}}^R_0$ via Eq.~\eqref{eq:tweedie}. The Object phase, conditioned on $\hat{\bm{\tau}}^R_0$, produces $\hat{\bm{\tau}}^O_0$ the same way, completing the differentiable chain $\bm{\tau}^R_t \to \hat{\bm{\tau}}^R_0 \to \hat{\bm{\tau}}^O_0 \to \mathcal{C}$. By contrast, the Joint denoiser produces both clean estimates in one pass from $(\bm{\tau}^R_t, \bm{\tau}^O_t)$, shortening the chain to $\bm{\tau}^R_t \to \hat{\bm{\tau}}^O_0 \to \mathcal{C}$.

\textbf{Guidance gradient.} The chain bypasses the intractability by substituting each Tweedie estimate for the corresponding clean trajectory throughout the derivation (Sec.~\ref{sec:guidance_derivation}). The DPS approximation~\cite{chungDiffusionPosteriorSampling2024} is applied once per phase, meaning Causal incurs two approximation errors against the Joint denoiser's one. We accept this so the object trajectory is generated as a response to the robot plan. Finally by chain rule,
\begin{equation}
\label{eq:guidance-grad-chain}
\nabla_{\bm{\tau}^R_t}\, \mathcal{C}
\;=\;
\underbrace{\frac{\partial \mathcal{C}}{\partial \hat{\bm{\tau}}^O_0}}_{\text{cost gradient}}
\;\cdot\;
\underbrace{\frac{\partial \hat{\bm{\tau}}^O_0}{\partial \hat{\bm{\tau}}^R_0}}_{\substack{\text{Object phase} \\ \text{backprop}}}
\;\cdot\;
\underbrace{\frac{\partial \hat{\bm{\tau}}^R_0}{\partial \bm{\tau}^R_t}}_{\substack{\text{Robot phase} \\ \text{backprop}}}
\end{equation}



\subsection{Guidance for the Causal Factorization}
\label{sec:guidance_derivation}

In this section, we derive the guidance formulation for the Causal factorization. Throughout this derivation, we apply two conditional independence assumptions: (i)~$\mathbf{y}$ depends on the true clean trajectories $(\bm{\tau}^R_0, \bm{\tau}^O_0)$, and (ii)~once the clean object trajectory is known, the robot trajectory carries little additional information about whether the goal was reached. The robot collision and retreat terms are the exception in $\mathcal{C}$, and we evaluate them on $\hat{\bm{\tau}}^R_0$ directly.

Applying total probability and assumption~(i), we expand $p(\mathbf{y} \mid \bm{\tau}^R_t)$ to include the clean robot trajectory, and substituting the DPS point estimate $\hat{\bm{\tau}}^R_0$ for the robot posterior gives $p(\mathbf{y} \mid \bm{\tau}^R_t) \approx p(\mathbf{y} \mid \hat{\bm{\tau}}^R_0)$. Applying total probability a second time to this allows us to introduce $\tau_{0}^{O}$, with assumption~(ii) dropping $\hat{\bm{\tau}}^R_0$ from a term. The Object phase produces $\hat{\bm{\tau}}^O_0$ from a noisy $\bm{\tau}^O_t$, so we expand $p(\bm{\tau}^O_0 \mid \hat{\bm{\tau}}^R_0)$ over $\bm{\tau}^O_t$ by total probability. Substituting this into the previous expression and swapping the order of integration yields
\begin{equation}
\label{eq:object_expectation}
p(\mathbf{y} \mid \hat{\bm{\tau}}^R_0)
=
\mathbb{E}_{\bm{\tau}^O_t \sim p(\bm{\tau}^O_t \mid \hat{\bm{\tau}}^R_0)}
\!\left[\, p(\mathbf{y} \mid \bm{\tau}^O_t, \hat{\bm{\tau}}^R_0) \,\right]
\end{equation}

Then, applying the DPS estimate once more within each draw gives $p(\mathbf{y} \mid \bm{\tau}^O_t, \hat{\bm{\tau}}^R_0) \approx p(\mathbf{y} \mid \hat{\bm{\tau}}^O_0(\bm{\tau}^O_t))$, where assumption~(ii) is applied to remove $\hat{\bm{\tau}}^R_0$. Since $\hat{\bm{\tau}}^O_0$ depends on $\bm{\tau}^O_t$, the integrand varies across draws.


\textbf{$K$-sample estimate.} While the expectation in \eqref{eq:object_expectation} is over $\bm{\tau}^O_t$, the current denoising step supplies only one draw. However, the Causal factorization lets us approximate draws from $p(\bm{\tau}^O_t \mid \hat{\bm{\tau}}^R_0)$ at the cost of an Object phase forward pass. We renoise the clean estimate $\hat{\bm{\tau}}^O_0$ to the current noise level $t$ with $K$ independent noise vectors and denoise each with the \emph{current} $\hat{\bm{\tau}}^R_0$ as conditioning, resulting in $K$ distinct predictions $\hat{\bm{\tau}}^{O,(k)}_0$ in one batched pass. We refer to this approach as Monte Carlo (MC) w/ DPS in Table~\ref{tab:main_results}.


Since the goal likelihood is a Boltzmann distribution, the estimate of \eqref{eq:object_expectation} is a mean of exponentials, and taking its log yields the guidance objective
\begin{equation}
\label{eq:free_energy}
{E}_\lambda
\;=\;
-\lambda \log\!\left[\frac{1}{K}\sum_{k=1}^{K}
\exp\!\left(-\,\mathcal{C}\big(\hat{\bm{\tau}}^{O,(k)}_0, \hat{\bm{\tau}}^R_0\big) / \lambda\right)\right]
\end{equation}

We scale the above by $\lambda$ to make the following gradient simpler: writing $\mathcal{C}_k = \mathcal{C}(\hat{\bm{\tau}}^{O,(k)}_0, \hat{\bm{\tau}}^R_0)$ and differentiating \eqref{eq:free_energy} gives $\partial {E}_\lambda / \partial \mathcal{C}_k = \mathrm{softmax}(-\mathcal{C}_k/\lambda)$ which we define as $w_k$. Then by the chain rule, where each $\nabla_{\bm{\tau}^R_t}\mathcal{C}_k$ follows the same differentiable chain as Eq.~\eqref{eq:guidance-grad-chain}: 
\begin{equation}
\label{eq:softmax_grad}
\nabla_{\bm{\tau}^R_t} {E}_\lambda
=
\sum_{k=1}^{K} w_k \, \nabla_{\bm{\tau}^R_t} \mathcal{C}_k,
\qquad
\sum_{k=1}^{K} w_k = 1
\end{equation}


\textbf{Softmax weighting.} A point estimate would report the cost of the mean predicted outcome, whereas \eqref{eq:free_energy} reports the cost over the distribution of possible outcomes. The two differ in our setting. For instance, the placement term $C_{\text{place}}$ penalizes overshoot more heavily than undershoot, so drawn trajectories that mostly undershoot with a few overshooting have a mean endpoint that escapes the penalty entirely. Evaluating $\mathcal{C}$ per draw avoids this, but leaves the question of how to combine the $K$ gradients. Because pushing is multimodal, averaging the $K$ draws yields a direction that suits no single outcome, so \eqref{eq:softmax_grad} weights toward the lowest-cost one instead. A Joint denoiser produces a single coupled $(\bm{\tau}^R, \bm{\tau}^O)$ pair per pass, whereas the factorization lets us draw $K$ object outcomes under one robot plan.



\subsection{Guidance cost design}
\label{sec:method:cost}

\textbf{Cost terms.} The five terms in $\mathcal{C}$ (Table~\ref{tab:cost-summary}) each contribute a distinct shaping signal. $C_{\text{place}}$ uses an asymmetric pinball function~\cite{koenker1978regression} to favor undershoot, which another push motion can correct. $C_{\text{pred}}$ adds an earlier signal so the planner can correct for a projected overshoot before $C_{\text{place}}$ would react. $C_{\text{retreat}}$ pulls the robot back from the object after a bad approach for the push. Bad approaches are rare in the expert demonstrations, so the planner has learned limited knowledge of retreat behavior. Without $C_{\text{retreat}}$, the robot could continue pushing and displace a placed object. 

\begin{table}[h]
\centering
\footnotesize
\setlength{\tabcolsep}{3pt}
\renewcommand{\arraystretch}{1.15}
\caption{Summary of the five guidance cost terms. Starred ($\star$) terms are novel contributions of this work, while the rest are inspired by~\cite{zhengDiffusionBasedPlanningAutonomous2025}. Each penalty is followed in \textit{italics} by the condition under which it is active.}
\label{tab:cost-summary}
\begin{tabular}{@{}l l >{\raggedright\arraybackslash}p{5.2cm}@{}}
\toprule
\textbf{Term} & \textbf{Role} & \textbf{Penalty} \\
\midrule
$C_{\text{place}}^{\star}$ & Push
  & Asymmetric endpoint error along the push axis (overshoot $=25\times$ undershoot) plus squared lateral error. \textit{Object endpoint near target.} \\
\addlinespace[2pt]
$C_{\text{pred}}^{\star}$ & Push
  & Penalizes the object's predicted landing position at the horizon end, assuming it keeps its current along-axis velocity. \textit{Lookahead to predict overshoot.} \\
\addlinespace[2pt]
$C_{\text{coll}}^{(o)}$ & Safety
  & Margin violation of the predicted object trajectory against placed objects and obstacles, through the barrier $\Psi(x){=}e^{x}{-}x{-}1$. \textit{Object penetrates margin.} \\
\addlinespace[2pt]
$C_{\text{coll}}^{(r)}$ & Safety
  & Same barrier on the predicted robot trajectory. \textit{Robot penetrates margin.} \\
\addlinespace[2pt]
$C_{\text{retreat}}^{\star}$ & Retreat
  & Penalizes robot proximity to the target object when retreat is needed. \textit{Bad approach, or target object is placed.} \\
\bottomrule
\end{tabular}
\end{table}

\textbf{Per-term gated step schedule.} At inference, we compute each cost term separately, averaging it over the $K$ draws with the weights $w_k$ of \eqref{eq:softmax_grad}. Any term with $\bar{\mathcal{C}}_i = \sum_k w_k \mathcal{C}_{i,k} < \eta$ is dropped to keep its noisy gradient from contributing to the step. We normalize each remaining weighted gradient to unit norm, scale it by a tuned factor $c_i$, and sum the results:
\begin{equation}
\label{eq:per-term-step}
\Delta = \sum_{i \,:\, \bar{\mathcal{C}}_i \geq \eta} c_i \cdot
\frac{\sum_{k} w_k \nabla_{\bm{\tau}^R_t} \mathcal{C}_{i,k}}
     {\big\|\sum_{k} w_k \nabla_{\bm{\tau}^R_t} \mathcal{C}_{i,k}\big\|_2 + \varepsilon}.
\end{equation} 
In Eq. \ref{eq:per-term-step}, we deviate from the residual-norm recipe of~\cite{chungDiffusionPosteriorSampling2024} since this scheme generalizes naturally to cost terms with different units and scales. For the Joint planner, $\mathcal{C}$ is evaluated on the single $\hat{\bm{\tau}}^O_0$ the pass produces, so Eq.~\eqref{eq:per-term-step} applies with $K=1$. The robot trajectory is then updated as $\bm{\tau}^R_{t-1} \leftarrow \mathrm{DDIM}(\bm{\tau}^R_t, \bm{v}_\theta) - \Delta$, applied at each denoising step within the guidance window.

\section{Experiments}
\label{sec:experiments}


In this section, we evaluate the Causal planner against baselines on non-prehensile tasks, starting with Push-T in simulation. We then run simulated and real-world multi-object pushing experiments with a mobile robot, applying the cost terms of Sec.~\ref{sec:method:cost} in both guidance and best-of-$N$ sampling.

\subsection{Push-T Experiments}
\paragraph{Task and Data} In Push-T~\cite{chiDiffusionPolicyVisuomotor2024}\cite{florenceImplicitBehavioralCloning2021}, the objective is to align the gray T-block to a green goal region by pushing with a blue cylindrical end effector (see Table~\ref{tab:pusht_results}). In the simulation experiments, the cylindrical pusher has a 2D state, while the block state is represented by eight keypoints, each with a 2D state. No image information is used as conditioning, and the goal pose is identical in every episode. The success criteria is defined as the T-block achieving at least 95\% intersection-over-union (IoU) within 300 environment steps. We refer to the IoU value as coverage throughout this work. The \textit{pusht\_keypoints} dataset from Hugging Face was used for training our models.

\paragraph{Metrics} SR denotes success rate, with Wilson 95\% confidence intervals shown below each entry. Max and Final are the mean per-episode maximum and terminal coverage, aggregated over all episodes regardless of success. Steps is the mean number of environment steps to completion, computed on successful episodes only, since failures always exhaust the $300$-step cap. The Motion Fidelity metrics are new to the Push-T setting. Contact consistency (CC) measures how often a predicted block trajectory moves despite the pusher not touching it. A step is contact-free if the pusher-to-block surface distance exceeds a small distance threshold, and is violated if the block's mean per-keypoint displacement over that step exceeds $2$ distance units. Object-motion error (OME) is the $\ell_2$ distance between the predicted and actual block-centroid displacement at the end of each executed action chunk.

\paragraph{Baselines and ablation design} Because the Causal factorization is designed to predict object motion, we first evaluate with best-of-$N$ sampling, selecting the robot trajectory whose predicted object trajectory ends closest to the goal. This defers Diffusion Policy~\cite{chiDiffusionPolicyVisuomotor2024} to Sec.~\ref{mobile_robot_sec}, since it denoises only the robot trajectory. We instead use Diffusion Forcing~\cite{chen2024diffusion}, which denoises both trajectories and emphasizes causality. We implement its transformer variant, moving the per-token noise schedule from the temporal axis to our two trajectories. Each object token is held at a noise level no lower than the robot tokens, so the object is never resolved ahead of the robot motion producing it.

\begin{table}[h]
\centering
\caption{Push-T simulation results with each method at its own best-of-$N$ draws per replan, selected by success rate over $N \in \{1, 8, 16, 32\}$: $N{=}16$ for Diffusion Forcing [100 DDIM] and Joint [100 DDIM], $N{=}32$ for Causal [100 DDIM], and $N{=}8$ for Causal [50 DDIM]. Causal consistently has the highest success rate across all batch sizes. Each row pools $75$ episodes ($25$ episodes $\times$ $3$ seeds: $42$, $1000$, $2000$). Max and Final are reported with their $95\%$ confidence intervals (CI), and SR with its Wilson $95\%$ CI. Latency is roughly constant across the different batch sizes for each model. All models have around $17.8$M parameters. Rows are color-coded: \colorbox{baselinerow}{\strut\,baselines\,}, \colorbox{ablationrow}{\strut\,ablations\,}, \colorbox{proposedrow}{\strut\,ours (proposed)\,}. \textbf{Bolded} entries mark the best per column.}
\label{tab:pusht_results}
\scriptsize
\setlength{\tabcolsep}{3pt}
\setlength{\extrarowheight}{0pt}
\renewcommand{\arraystretch}{1.2}
\newcommand{\stk}[2]{\shortstack{\rule{0pt}{2.6ex}#1\\[-1pt]{\tiny #2}\rule[-1.0ex]{0pt}{0pt}}}
\begin{tabular}{|l|cc|cc|cc|}
\hline
\multicolumn{1}{|c|}{\multirow{4}{*}{\includegraphics[height=1.1cm]{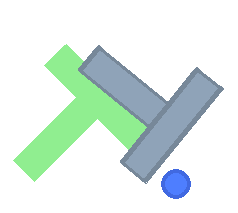}}}
  & \multicolumn{6}{c|}{\textbf{Push-T}} \\
\cline{2-7}
  & \multicolumn{2}{c|}{\textbf{Coverage}}
  & \multicolumn{2}{c|}{\textbf{Task}}
  & \multicolumn{2}{c|}{\textbf{Motion Fidelity}} \\
\cline{2-7}
  & Max & Final & SR & Steps & CC & OME \\
  & (\%)$\uparrow$ & (\%)$\uparrow$ & (\%)$\uparrow$ & $\downarrow$ & (\%)$\downarrow$ & $\downarrow$ \\
\hline
\rowcolor{baselinerow} Diffusion Forcing
  & \stk{91.5}{$\pm$2.5}
  & \stk{85.3}{$\pm$3.4}
  & \stk{44.0}{[33.3,\,55.3]}
  & 201.4 & 29.3 & \textbf{2.64} \\
\hline
\rowcolor{ablationrow} Joint
  & \stk{92.5}{$\pm$1.3}
  & \stk{87.3}{$\pm$2.5}
  & \stk{40.0}{[29.7,\,51.3]}
  & 213.1 & 18.3 & 3.91 \\
\hline
\rowcolor{proposedrow} Causal [100 DDIM]
  & \stk{\textbf{94.1}}{$\pm$0.7}
  & \stk{\textbf{91.6}}{$\pm$1.5}
  & \stk{\textbf{54.7}}{[43.4,\,65.4]}
  & 190.5 & 11.6 & 3.06 \\
\hline
\rowcolor{proposedrow} Causal [50 DDIM]
  & \stk{92.9}{$\pm$2.6}
  & \stk{89.9}{$\pm$2.9}
  & \stk{53.3}{[42.2,\,64.2]}
  & \textbf{174.7} & \textbf{10.8} & 2.96 \\
\hline
\end{tabular}
\end{table}


\paragraph{Sim results}
Table~\ref{tab:pusht_results} reports each method at its best sampling budget. Causal attains the highest success rate while completing in the fewest steps, with tighter confidence intervals on coverage. The sharpest separation, however, is contact consistency. Causal predicts uncaused object motion on $11.6\%$ of contact-free steps, against $18.3\%$ and $29.3\%$ for the baselines, and this ordering holds at every batch size we evaluated. Alongside each rollout, we also log the standard deviation across the $N$ draws for the predicted object trajectories. The object spread orders the success rate across all three methods ($2.79$, $2.98$, $3.71$ for Causal, Diffusion Forcing, and Joint). When the $N$ draws disagree, the $\arg\min$ selects the most optimistic object prediction, which may be physically implausible given the robot trajectory. When the two trajectories are consistent, it selects on plan quality. This explains why Causal is the only method that improves from $N{=}16$ to $N{=}32$, gaining $10.7$ percentage points in SR while Diffusion Forcing and Joint lose $10.7$ and $17.3$.

We attribute Causal's performance improvement to two changes: $v$-prediction, and the gate blending the clean robot estimate tokens toward a learned null embedding at high diffusion timesteps (Sec.~\ref{sec:method:arch}). In preliminary experiments ablating each factor over $30$ episodes per condition, success rate improves from $16.7\%$ ($\epsilon$-prediction, no gate) to $33.3\%$ with $v$-prediction alone, $40.0\%$ with the gate alone, and $56.7\%$ with both.

We report latency as the wall-clock time for one replan. To roughly match latency across the three methods in Table~\ref{tab:pusht_results}, we also report the Causal planner at 50 DDIM steps, which still outperforms the baselines on nearly every metric. Its median latency is 2008~ms, against around 2300~ms for the baselines and 4257~ms for Causal at 100 DDIM steps. Lastly as a reference upper bound, the official leRobot Diffusion Policy for PushT, which uses the same evaluation environment as we do, reports a 65\% task success rate and 95.5\% mean max coverage over 500 episodes. However, this is with a model that includes image observations and has 0.3B parameters. Causal achieves close to similar results with 17$\times$ fewer parameters.

\subsection{Mobile Robot Push Experiments}
\label{mobile_robot_sec}
\paragraph{Task and data} Our scenario is a mobile robot that stages $M$ boxes at target configurations $\mathcal{T}_1, \dots, \mathcal{T}_M$. We model a deployment where each box is handed off to downstream automation the moment it's placed. Bumping it afterward would be unsafe and stall the pipeline, so we end the trial as soon as any placed box is disturbed. Boxes are 8~cm per side, and each target is the box footprint inflated by a tolerance $\delta \in \{4, 2\}$ cm. Training data consists of roughly 500 human teleoperation episodes at $\delta = 2$ cm, with any episode that bumped a placed box discarded. We evaluate the planners on three environments that share structural elements with the training data, but never appear in their exact configurations. We use the TSP planner to assign boxes to subtasks throughout the experiments. Lastly, all diffusion models are only given a local ego-centric crop when it comes to info on nearby objects.


\paragraph{Metrics} Our primary success metric is the \emph{Box Placement Fraction}, or BPF. It records the average fraction of boxes placed within their target configurations across all episodes. We prefer this fractional measure over binary episode success because trials end the instant any placed box is disturbed, a rule that mirrors the real-world requirement that each box remain at its target after placement. Binary success would collapse 7 of 8 placed and 0 of 8 placed into the same zero, while BPF preserves credit for placements completed before the failure. We also report \emph{Shortest Path Length}, abbreviated SPL and adapted from~\cite{andersonEvaluationEmbodiedNavigation2018}. We compute it per box rather than per episode as $\mathrm{SPL} = \tfrac{1}{K}\sum_{i=1}^{K} S_i \, \ell_i^{\star} / \max(p_i, \ell_i^{\star})$, where $K$ is the total number of placement attempts across all evaluation episodes, $S_i \in \{0,1\}$ indicates whether placement $i$ succeeded, $\ell_i^{\star}$ is the shortest feasible path that completes placement $i$, and $p_i$ is the path the planner took.

\paragraph{Baselines and ablation design} Our main baseline is Diffusion Policy~\cite{chiDiffusionPolicyVisuomotor2024}, an established baseline for pushing tasks and one that denoises only the robot trajectory. Since guidance depends on object motion fidelity, we use the joint planner rather than Diffusion Forcing~\cite{chen2024diffusion}. The joint planner attains better contact consistency, and the OME gap of roughly one distance unit is small relative to the Push-T workspace scale (Table~\ref{tab:pusht_results}). In addition, we evaluated a cross-entropy method (CEM) baseline optimizing the same guidance cost (Sec.~\ref{sec:method:cost}), using the simulator as a forward model and rolling out $64\;\text{samples}\times5\;\text{iters}=320$ candidates per replan. Our costs give no signal for approach, so a deterministic guide policy handles it, and it uses the true global occupancy map rather than the egocentric crop all diffusion methods see. The ablations instead focus on the diffusion planners, isolating two factors. DPS vs.\ Best-of-N compares the two ways of applying the cost in Sec.~\ref{sec:method:cost}, as a gradient during denoising or as an objective to minimize over 16 sampled plans. Joint vs.\ Causal isolates the architecture and tests whether the causal factorization itself contributes to the gain. Lastly, all diffusion-based planners have roughly $75$--$90$~M parameters.


\begin{table*}[h]
\centering
\caption{Simulation results across 3 previously unseen environments at two goal tolerances ($4$ cm, $2$ cm), pooled over $150$ episodes per cell ($3$ seeds $\times$ $50$ episodes). All diffusion methods used 100 DDIM steps, unless otherwise stated, and are trained at a tolerance of $2$ cm. Diffusion Policy uses FiLM~\cite{perez2018film} for conditioning whereas Joint and Causal use AdaLN~\cite{peeblesScalableDiffusionModels2023}. For best of N sampling, the sample that minimizes the cost function in Sec.~\ref{sec:method:cost} is selected. SPL and BPF are both reported out of $100$ and with their $95\%$ CIs. \textbf{Bolded} entries mark the best BPF per (environment, tolerance). Causal w/ MC DPS is run at $K=16$. Rows are color-coded: \colorbox{baselinerow}{\strut\,baselines\,}, \colorbox{ablationrow}{\strut\,ablations\,}, \colorbox{proposedrow}{\strut\,ours (proposed)\,}.}
\label{tab:main_results}
\setlength{\tabcolsep}{3pt}
\renewcommand{\arraystretch}{1.1}
\setlength{\aboverulesep}{0pt}
\setlength{\belowrulesep}{0pt}
\setlength{\extrarowheight}{0pt}
\begin{tabular}{@{}p{3.2cm}|>{\centering\arraybackslash}p{1.0cm}|c|*{2}
{>{\centering\arraybackslash}p{1.6cm}}|*{2}{>{\centering\arraybackslash}p{1.6cm}}|*{2}{>{\centering\arraybackslash}p{1.6cm}}@{}}
\multicolumn{3}{c}{} & \multicolumn{2}{c|}{\textbf{U-Shape}} & \multicolumn{2}{c|}{\textbf{Pallet Staging}} & \multicolumn{2}{c}{\textbf{Cross-Dock Staging}} \\
\multicolumn{3}{c}{} & \multicolumn{2}{c|}{\includegraphics[width=3cm]{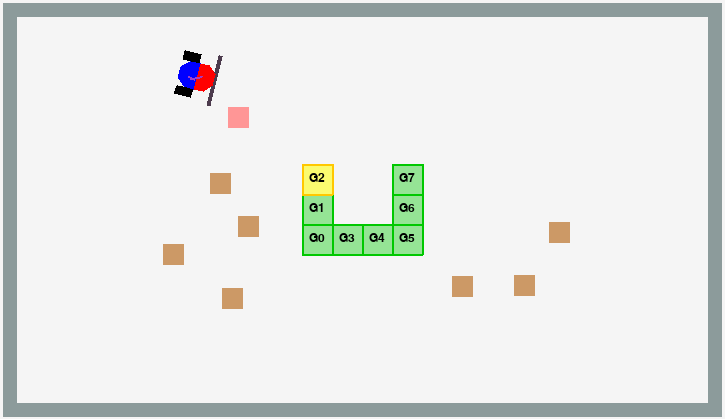}} & \multicolumn{2}{c|}{\includegraphics[width=3cm]{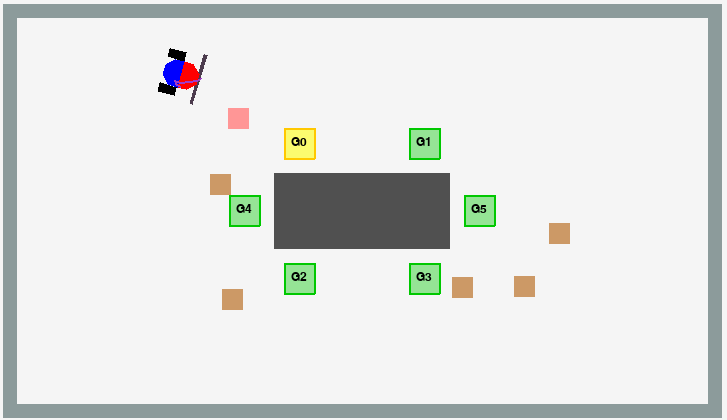}} & \multicolumn{2}{c}{\includegraphics[width=3cm]{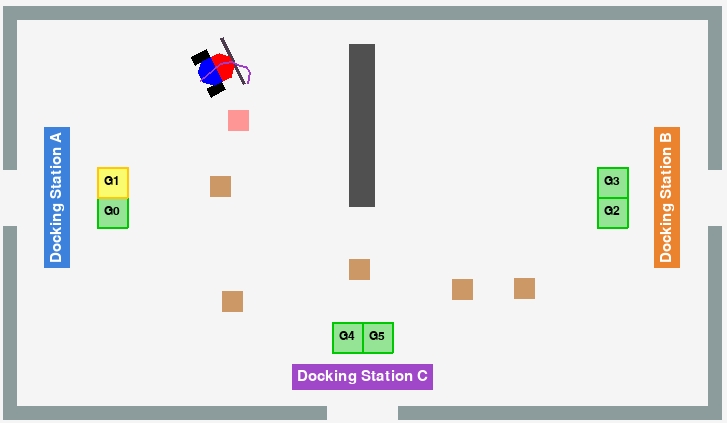}} \\
\cmidrule(lr){4-5}\cmidrule(lr){6-7}\cmidrule(lr){8-9}
\multicolumn{1}{c}{\textbf{Method}} & \multicolumn{1}{c}{\shortstack{\textbf{Lat.}\\(ms)$\downarrow$}} & \multicolumn{1}{c}{\shortstack{\textbf{Tol.}\\(cm)}} & SPL$\uparrow$ & BPF$\uparrow$ & SPL$\uparrow$ & BPF$\uparrow$ & SPL$\uparrow$ & BPF$\uparrow$ \\
\midrule

\rowcolor{baselinerow}  & & 4 & 61.7{\scriptsize$\,\pm$2.6} & 26.1{\scriptsize$\,\pm$2.8} & 67.2{\scriptsize$\,\pm$2.4} & 36.6{\scriptsize$\,\pm$3.5} & 68.0{\scriptsize$\,\pm$2.3} & 40.6{\scriptsize$\,\pm$4.4} \\
\rowcolor{baselinerow} \multirow{-2}{=}{\makecell[l]{CEM\\(oracle dynamics)}} & \multirow{-2}{*}{2209} & 2 & 57.6{\scriptsize$\,\pm$3.2} & 19.2{\scriptsize$\,\pm$1.9} & 64.3{\scriptsize$\,\pm$2.7} & 33.1{\scriptsize$\,\pm$3.6} & 64.8{\scriptsize$\,\pm$2.7} & 29.8{\scriptsize$\,\pm$2.9} \\
\midrule

\rowcolor{baselinerow}  & & 4 & 46.7{\scriptsize$\,\pm$1.8} & 38.3{\scriptsize$\,\pm$4.0} & 48.1{\scriptsize$\,\pm$2.0} & 48.0{\scriptsize$\,\pm$4.8} & 47.9{\scriptsize$\,\pm$2.2} & 47.1{\scriptsize$\,\pm$5.3} \\
\rowcolor{baselinerow} \multirow{-2}{=}{Diffusion Policy} & \multirow{-2}{*}{716} & 2 & 35.3{\scriptsize$\,\pm$2.2} & 21.8{\scriptsize$\,\pm$2.1} & 34.3{\scriptsize$\,\pm$2.6} & 33.7{\scriptsize$\,\pm$3.7} & 34.2{\scriptsize$\,\pm$2.5} & 29.8{\scriptsize$\,\pm$3.8} \\
\midrule

\rowcolor{ablationrow}  & & 4 & 50.3{\scriptsize$\,\pm$1.8} & 41.7{\scriptsize$\,\pm$4.5} & 54.3{\scriptsize$\,\pm$1.7} & 54.7{\scriptsize$\,\pm$5.1} & 52.6{\scriptsize$\,\pm$2.0} & 55.4{\scriptsize$\,\pm$5.3} \\
\rowcolor{ablationrow} \multirow{-2}{=}{Joint w/ DPS} & \multirow{-2}{*}{1893} & 2 & 46.9{\scriptsize$\,\pm$2.0} & 31.0{\scriptsize$\,\pm$3.5} & 49.7{\scriptsize$\,\pm$1.9} & 51.7{\scriptsize$\,\pm$5.1} & 50.1{\scriptsize$\,\pm$2.3} & 44.6{\scriptsize$\,\pm$4.7} \\
\midrule

\rowcolor{ablationrow}  & & 4 & 50.9{\scriptsize$\,\pm$1.4} & \textbf{54.9}{\scriptsize$\,\pm$5.2} & 56.7{\scriptsize$\,\pm$1.4} & \textbf{68.8}{\scriptsize$\,\pm$5.4} & 54.2{\scriptsize$\,\pm$1.8} & 62.8{\scriptsize$\,\pm$5.5} \\
\rowcolor{ablationrow} \multirow{-2}{=}{Joint Best-of-16 Sampling} & \multirow{-2}{*}{5088} & 2 & 48.0{\scriptsize$\,\pm$1.7} & 38.9{\scriptsize$\,\pm$4.3} & 54.9{\scriptsize$\,\pm$1.5} & \textbf{61.8}{\scriptsize$\,\pm$5.5} & 52.4{\scriptsize$\,\pm$1.9} & 55.8{\scriptsize$\,\pm$5.3} \\
\midrule

\rowcolor{proposedrow}  & & 4 & 49.7{\scriptsize$\,\pm$1.9} & 40.4{\scriptsize$\,\pm$4.4} & 52.9{\scriptsize$\,\pm$1.9} & 49.7{\scriptsize$\,\pm$5.3} & 50.9{\scriptsize$\,\pm$2.1} & 56.2{\scriptsize$\,\pm$5.2} \\
\rowcolor{proposedrow} \multirow{-2}{=}{Causal w/ MC DPS} & \multirow{-2}{*}{4163} & 2 & 49.4{\scriptsize$\,\pm$2.5} & 26.7{\scriptsize$\,\pm$3.2} & 50.7{\scriptsize$\,\pm$2.1} & 45.9{\scriptsize$\,\pm$4.8} & 49.0{\scriptsize$\,\pm$2.2} & 48.6{\scriptsize$\,\pm$4.7} \\
\midrule

\rowcolor{proposedrow}  & & 4 & 53.2{\scriptsize$\,\pm$2.2} & 52.8{\scriptsize$\,\pm$5.0} & 57.4{\scriptsize$\,\pm$1.9} & 66.6{\scriptsize$\,\pm$5.3} & 54.7{\scriptsize$\,\pm$2.3} & \textbf{65.7}{\scriptsize$\,\pm$5.1} \\
\rowcolor{proposedrow} \multirow{-2}{=}{Causal Best-of-16 Sampling [50 DDIM]} & \multirow{-2}{*}{2812} & 2 & 53.2{\scriptsize$\,\pm$2.5} & \textbf{40.8}{\scriptsize$\,\pm$4.6} & 55.6{\scriptsize$\,\pm$2.4} & 60.8{\scriptsize$\,\pm$5.4} & 54.4{\scriptsize$\,\pm$2.5} & \textbf{60.7}{\scriptsize$\,\pm$5.2} \\

\bottomrule
\end{tabular}
\end{table*}


\paragraph{Sim results} Table~\ref{tab:main_results} reports BPF and SPL across three environments, each chosen to probe a different failure mode. \emph{U-Shape} tests path feasibility, since placed boxes block routes across the U and force the robot to reroute. \emph{Cross-Dock Staging} tests planning over long horizons. \emph{Pallet Staging} tests overshoot in isolation, since any overshoot wedges the current box against the central pillar. Across the table, all cost-driven diffusion methods outperform the baselines in all trials. Best-of-16 sampling outperforms DPS guidance for both architectures, and the two Best-of-16 variants are comparable, with BPF confidence intervals (CIs) overlapping in every cell of Table~\ref{tab:main_results}. This gap therefore reflects how the cost is applied rather than the architecture, since guidance is bound to its initial sample while Best-of-16 is not. The separation between Causal and Joint is in efficiency and handling tighter tolerances. At 2 cm, Causal Best-of-16 attains the best placement of any method in two of the three environments, though the margin over Joint falls within our CIs. It also has a higher SPL than Joint Best-of-16 in all six cells, with CIs separating only on U-Shape at $2$~cm. These results are all while Causal runs half the denoising steps, 50 against Joint's 100. In preliminary experiments, Joint performed slightly better at 100 steps than at 50, whereas Causal was comparable at both, matching the trend in the Push-T results (Table~\ref{tab:pusht_results}). When comparing Joint and Causal with DPS guidance, we observe similar performance, as the BPF CIs all overlap. Causal w/ MC DPS separates on efficiency in one environment, completing the long-horizon Cross-Dock task in $4.8\%$ fewer environment steps than guided Joint at $4$~cm and $12.2\%$ fewer at $2$~cm. However, the clearest separation between the two methods is latency, which we examine next.


\paragraph{Latency} The Causal factorization uses two decoder passes per denoising step, so the DDIM reduction makes the two methods match at 100 decoder passes per replan. However, when averaged across environments and tolerances, Causal Best-of-16 replans in $2812$\,ms against Joint's $5088$\,ms, a $1.8\times$ reduction. Two mechanisms produce this gap. Causal skips object denoising on non-pushing replans (i.e., navigation), so $49\%$ of its replans use only 50 decoder passes and cost $1812$\,ms, rather than the $3776$\,ms a replan costs when it denoises an object as well. Both methods run 100 decoder passes on a replan that requires denoising an object, but not the same passes. The first pass of each Causal denoising step runs over the robot trajectory alone. As a result, this pass skips the spatial self-attention block and halves the query length of the cross-attention blocks. In contrast, every Joint denoising step runs on both trajectories, so Joint never gets a cheaper pass. Causal's two passes per denoising step cost $75$\,ms against Joint's $51$\,ms, but it wins on total replan cost by running half as many steps.



\paragraph{Guidance versus conditioning}\label{par:guidance-vs-cond} This is an important comparison, and it is reported in Table~\ref{tab:main_results}, where guidance (Causal w/ MC DPS) gives an 8.1-point BPF gain on average and up to 18.8 points over Diffusion Policy, an unguided baseline. All models condition on scene occupancy, placed boxes, goal pose, and push direction via cross-attention, and train \emph{only} on demonstrations in which no box was overshot. Yet, the unguided planners still overshoot, and we demonstrate this in a plan-only test. In each of $50$ Pallet Staging episodes, we pause at the last replan step that still determines the first box placement, draw $50$ plans from that identical state, and record the predicted along-axis placement error $e_\parallel$ at the end of the horizon. Despite no overshooting trajectory in training, $3.6\%$ of the $2500$ unguided plans land more than $1$~cm past the goal and $0.64\%$ more than $2$~cm. At identical state and initial noise, guidance drops these to $2.6\%$ and $0.20\%$, respectively. 



\paragraph{OOD Robot Demo} We recreate a portion of the Cross-Dock Staging environment from Table~\ref{tab:main_results} and deploy the Causal w/ MC DPS model without fine-tuning it (Fig.~\ref{fig:robot_demo}). An overhead camera tracks AprilTags on the robot and each box at 20\,Hz, and a low-level pose controller follows the planner's predicted trajectory at 30\,Hz. We ran $10$ trials at a 3 cm tolerance for each Causal w/ MC DPS and Diffusion Policy (DP) on four new objects of differing mass and size: a triangular prism, cylinder, pentagon, and star (Fig.~\ref{fig:robot_demo}). Causal w/ MC DPS achieves an $85.0$ mean BPF with $60\%$ of episodes fully completed, compared to $37.5$ and $0\%$ for DP. Our guidance terms use only the object's center and the push axis, so guidance transfers to new shapes unchanged while the prior alone does not. These improved rates also benefit from only four goal slots, leaving a less constrained c-space and a lower chance of displacing prior-placed objects.



\begin{figure}[t]
\centering
\begin{subfigure}[t]{0.48\columnwidth}
  \centering
  \includegraphics[width=\linewidth]{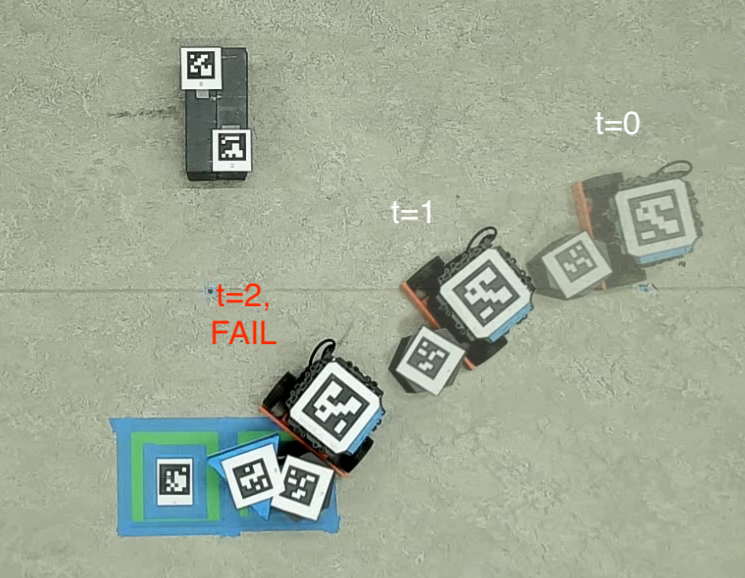}
  \caption{Pushed box hits a prior placement, failing the episode.}
  \label{fig:box_fail}
\end{subfigure}
\hfill
\begin{subfigure}[t]{0.48\columnwidth}
  \centering
  \includegraphics[width=\linewidth]{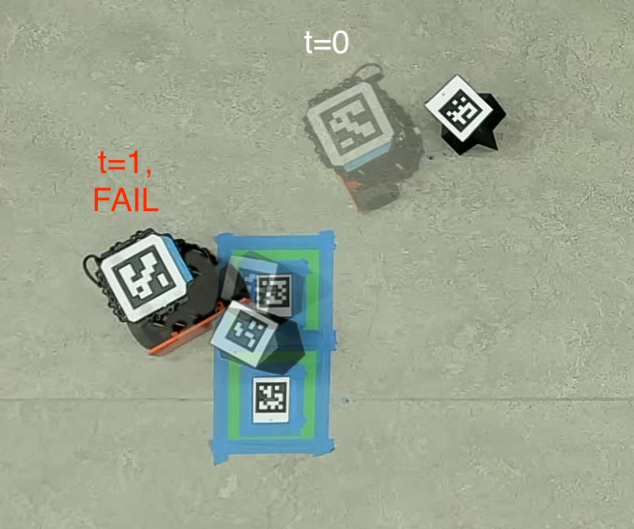}
  \caption{The robot hits a prior placement while navigating, failing the episode.}
  \label{fig:robot_fail}
\end{subfigure}
\caption{Failure modes.}
\vspace{-0.5cm}
\label{fig:failures}
\end{figure}

\paragraph{Failure Modes} We bin each episode into five outcomes: success, a placed box disturbed by the pushed box (or the robot), a timeout with the target wedged (or still moving). For the Best-of-16 planners, the top failure mode is disturbed by the pushed box, at around 33\% of all episodes for both Causal and Joint (see Fig.~\ref{fig:box_fail}). For the guidance methods and baselines, the top failure mode is robot disturbance, accounting for roughly 45\% of all episodes (see Fig.~\ref{fig:robot_fail}). 


\paragraph{Limitations and Future Work}
Our hardware demonstration relies on an overhead camera tracking AprilTags on the robot and each box, which supplies the global pose estimates that the planner conditions on. An onboard camera deployment is not a drop-in replacement. During a push, the box occludes the goal tag on the floor, so the robot loses sight of the goal precisely when the stopping decision must be made. Closing this gap is the goal of future work. In addition, the binary success rates are low across Table~\ref{tab:main_results}, ranging from $7\%$ to $37\%$ for Causal Best-of-16. Full task success requires placing every box within tolerance without disturbing any earlier placement, and a trial is considered a failure the instant one is disturbed. The difficulty also compounds with each placement, as every box placed narrows the c-space for the robot.



\section{Conclusion}

We formalized multi-object pushing as a multi-target assignment problem for a mobile robot, and proposed a diffusion-based planner with a Causal factorization: each robot denoising step is followed by an object step conditioned on the current clean robot estimate. This makes the predicted object motion a response to the robot plan, and exposes the object trajectory as a conditional that we can sample repeatedly. On Push-T, our model attains the highest success rate, $10.7$ percentage points over the closest baseline, while having the lowest rate of implausible object motion. Across three new multi-box pushing environments, Causal best-of-16 lifts placement by $21.5$ BPF points on average and up to $30.9$ over an unguided baseline, while matching a joint denoiser on placement performance and replanning $1.8\times$ faster. Lastly, on a TurtleBot 3 Burger, our guided Causal planner places $85.0\%$ of objects with geometries absent from training, against $37.5\%$ for Diffusion Policy.




\bibliographystyle{IEEEtran}
\bibliography{bib_Diffusion_Precision_Pusher}

\end{document}